\documentclass{article}

\PassOptionsToPackage{numbers, compress}{natbib}

\usepackage[preprint]{neurips_2022}

\usepackage[utf8]{inputenc} 
\usepackage[T1]{fontenc}    
\usepackage{hyperref}       
\usepackage{url}            
\usepackage{booktabs}       
\usepackage{amsfonts}       
\usepackage{nicefrac}       
\usepackage{microtype}      
\usepackage{xcolor}         

\usepackage{microtype}
\usepackage[pdftex]{graphicx}
\usepackage{subfigure}
\usepackage{booktabs} 

\usepackage{amsmath}
\usepackage{amsthm}
\usepackage{amssymb}

\usepackage{algorithmic}
\usepackage{threeparttable}
\usepackage{multirow}

\usepackage[ruled,vlined,linesnumbered]{algorithm2e}
\SetKwComment{Comment}{$\triangleright$\ }{}

\def\eqref#1{Equation~\ref{#1}}			

\title{EventMix: An Efficient Augmentation Strategy \\ for Event-Based Data}

\author{%
	{Guobin Shen$^{1, 4}$, \ Dongcheng Zhao$^{1}$, \ Yi Zeng$^{1, 2, 3, 4, 5}$}\thanks{Corresponding Author.}\\
	$^1$ Research Center for Brain-Inspired Intelligence, CASIA\\ 
      $^2$ Center for Excellence in Brain Science and Intelligence Technology, CAS\\
      $^3$ National Laboratory of Pattern Recognition, CASIA\\
      $^4$ School of Future Technology, University of Chinese Academy of Sciences \\
      $^5$ School of Artificial Intelligence, University of Chinese Academy of Sciences \\
	\texttt{\{shenguobin2021, zhaodongcheng2016, yi.zeng\}@ia.ac.cn}
}

\begin{document}

\maketitle

\begin{abstract}
      High-quality and challenging event stream datasets play an important role in the design of an efficient event-driven mechanism that mimics the brain. Although event cameras can provide high dynamic range and low-energy event stream data, the scale is smaller and more difficult to obtain than traditional frame-based data, which restricts the development of neuromorphic computing. Data augmentation can improve the quantity and quality of the original data by processing more representations from the original data. This paper proposes an efficient data augmentation strategy for event stream data: $\textbf{EventMix}$. We carefully design the mixing of different event streams by Gaussian Mixture Model to generate random 3D masks and achieve arbitrary shape mixing of event streams in the spatio-temporal dimension. By computing the relative distances of event streams, we propose a more reasonable way to assign labels to the mixed samples. The experimental results on multiple neuromorphic datasets have shown that our strategy can improve its performance on neuromorphic datasets both for ANNs and SNNs, and we have achieved state-of-the-art performance on DVS-CIFAR10, N-Caltech101, N-CARS, and DVS-Gesture datasets.
\end{abstract}

\section{Introduction}
The event camera, such as the Dynamic Vision Sensor (DVS), is a bionic vision sensor that mimics how the human retina works. In contrast to the conventional cameras, the intensity change of each pixel is recorded asynchronously in an event-driven manner, rather than capturing intensity images at a fixed rate. Because of the high temporal resolution, high dynamic range,  low time latency, and energy consumption of event cameras, they are widely used in several domains, such as image reconstruction~\citep{zou2021learning,zhao2021spk2imgnet}, flow estimation~\citep{zhu2018ev,zhu2019unsupervised}, motion segmentation~\citep{stoffregen2019event}, and recognition~\citep{amir2017low,liCIFAR10DVSEventStreamDataset2017}, which also promotes the construction of frame-based datasets.

\begin{table*}[h!]
      \centering
      \begin{threeparttable}[b]
            \caption{The characteristic of the famous existing datasets.}
            \begin{tabular}{ccccc}
                  \toprule[1pt]
                  Dataset                                                 & Categories & Samples & Access                \\
                  \hline
                  DVS-CIFAR10~\cite{liCIFAR10DVSEventStreamDataset2017}   & 10         & 10,000  & Shot on static images \\
                  DVS-Gesture~\cite{amirLowPowerFully2017}                & 11         & 1342    & Shot on real scenes   \\
                  N-MNIST~\cite{orchardConvertingStaticImage2015}         & 10         & 70000   & Shot on static images \\
                  N-Caltech101~\cite{orchardConvertingStaticImage2015}    & 101        & 8709    & Shot on static images \\
                  N-CARS~\cite{sironiHATSHistogramsAveraged2018a}         & 2          & 24029   & Shot on real scenes   \\
                  N-Omniglot~\cite{liNOmniglotLargescaleNeuromorphic2021} & 1623       & 32460   & Shot on static images \\
                  \bottomrule[1pt]
            \end{tabular}

            \label{dataset}
      \end{threeparttable}
\end{table*}

As can be seen in Table~\ref{dataset}, compared to the traditional frame-based datasets, event-based datasets are smaller in scale, and the DVS-Gesture has only 1342 samples for 11 categories. Such small and sparse datasets can easily lead to overfitting and unstable convergence, whether for artificial neural networks (ANNs) or spiking neural networks (SNNs). Moreover, it has restricted the development of the event-based algorithm. An intuitive idea is to collect more event-based data. However, due to the scarcity of event cameras, the cost is expensive compared to collecting traditional frame-based datasets. An alternative approach is to apply the data augmentation to the existing datasets. The data augmentation approach improves the quality and quantity of the training data by adding prior knowledge such as rotation and flipping to generate more different representations of the training data. Researchers have proposed many strategies~\cite{shortenSurveyImageData2019, antoniouDataAugmentationGenerative2018, limFastAutoAugment2019, cubukRandAugmentPracticalAutomated2019b} for the traditional image data.

\begin{figure}[h!]
      \centering
      \includegraphics[width=13.5cm]{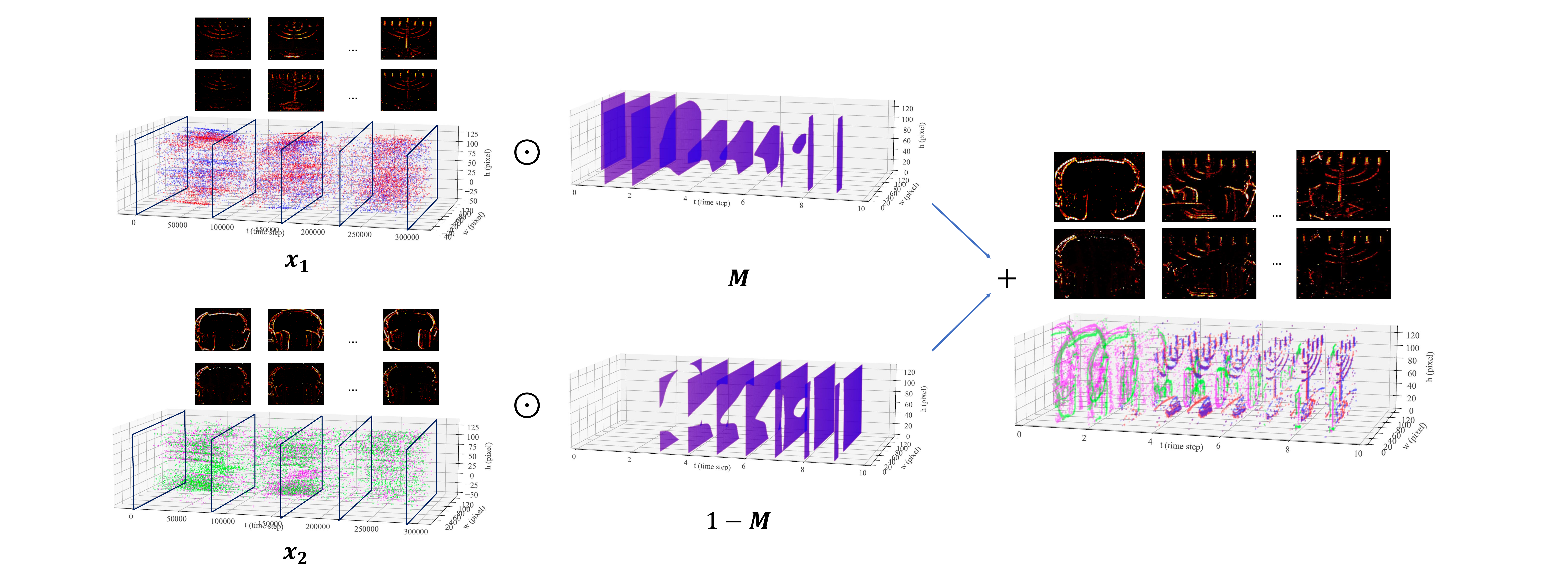}
      \caption{The illustration of our EventMix data augmentation strategy on events data.}
      \label{overview}
\end{figure}

However, these strategies do not take the sparse and spatio-temporal dynamic nature of the event stream data and cannot be directly applied. This paper proposes the $\textbf{EventMix}$ strategy, as shown in Figure~\ref{overview}, which fully takes the sample mixing and the characteristics of event stream data into consideration. The experimental results on several famous datasets have demonstrated that our EventMix is an efficient augmentation strategy for event-stream data. Our contributions are summarized as follows:

\begin{itemize}
      \item By thinking about the spatio-temporal dynamic of event stream data, we propose a random 3D mask generation method, which can generate masks of event stream data in temporal and spatial dimensions and provide a diverse way of mixing data stream data.
      \item We propose two mixing methods for event stream data labels. By calculating the relative distance between the original samples and the mixed samples, or the number of events in the mixed region, we achieve a reasonable label weight assignment for the event stream data labels.
      \item We validate the proposed EventMix event augmentation strategy on the DVS-CIFAR10~\cite{liCIFAR10DVSEventStreamDataset2017}, N-Caltech101~\cite{orchardConvertingStaticImage2015} , N-Cars~\cite{sironiHATSHistogramsAveraged2018}, and DVS-Gesture~\cite{amirLowPowerFully2017} datasets. The experimental results show that the proposed EventMix can effectively augment the event stream data and achieve state-of-the-art.
\end{itemize}

\section{Related Work}

\subsection{Event Stream Data}

Unlike conventional image data that describe the environment by the brightness and color information, event stream data describe the visual information of the environment by the change of brightness. The time stream information consists of four dimensions $(t, x, y, p)$. Where $t$ is the time of the event. $(x, y)$ are the coordinates of the pixel that perceived the event. $p$ is the polarity of the event. When the pixel perceives a brightness enhancement above a certain threshold, $+1$ is output; on the contrary, if the pixel perceives a brightness reduction above the threshold, $-1$ is output; otherwise, $0$ is output. Event stream data is widely used in neuromorphic computing~\cite{diehlUnsupervisedLearningDigit2015a, zhangSpikeTrainLevelBackpropagation2019a, lianTrainingDeepConvolutional2022, zhaoGLSNNMultiLayerSpiking2020, wuSpatioTemporalBackpropagationTraining2018} because it is similar to the biological retina in that it describes the environment by events that are similar to spikes.

With the development of deep learning, event stream data with low energy consumption, low latency, and biological plausibility has also gained attention. Gehrig et al.~\cite{gehrigEndtoEndLearningRepresentations2019} proposed a general framework that can process event stream data and output image-like, or video-like data, enabling neural networks designed for images and videos to be easily applied to event stream data. Schaefer et al.~\cite{schaeferAEGNNAsynchronousEventbased} designed an asynchronous event-based graph neural network that can process event stream data directly. However, its performance is not yet comparable to other frame-based methods. Therefore, in this study, we use the same method as Gehrig et al. for event stream data processing.

\subsection{Data Augmentation}

It is widely believed that a larger number of training samples can reduce the overfitting of neural networks and enhance the generalization ability of the model and its adaptability to new samples. However, large data sets often require significant consumption. Data augmentation is a common technique in deep learning to generate virtual samples around the training samples through prior knowledge to diversify the datasets and make the trained model more generalizable. The augmentation methods for image data have been well studied. For example, morphological transformation of images, such as random flip, random rotation and random crop~\cite{krizhevskyImageNetClassificationDeep2017a}; transformation of image pixels, such as adding noise~\cite{bishopTrainingNoiseEquivalent1995}, blurring, erasing, etc.

Some data augmentation methods have been proposed to combine multiple samples to generate new samples in recent years. For example, MixUp~\cite{zhangMixupEmpiricalRisk2017} combines different samples convexly and performs the same operation on the corresponding labels. CutMix~\cite{yunCutMixRegularizationStrategy2019} combines different regions of image samples and mixes the corresponding labels according to the area of different samples in the mixed sample. Puzzle Mix~\cite{kimPuzzleMixExploiting2020} mixes the computed salient regions of different samples to generate more efficient mixing instances. These mixing-based methods are believed to lead the models to empirical risk minimization estimates~\cite{carratinoMixupRegularization2020}, enabling many advanced deep neural network models to further improve their performance. However, since these methods are designed for image data, it is difficult to apply them directly to event stream data to achieve desired results.

Image data augmentation strategies have improved deep learning performance in tasks such as image classification. However, augmentation strategies for event stream data have not received sufficient attention, which also limits the further improvement of the performance of neural networks processing event streams. To the best of the authors' knowledge, only limited research has been conducted on augmenting event stream data. EventDrop~\cite{guEventDropDataAugmentation} achieves augmentation of event stream data by randomly dropping events, but they do not take into account the relationship between different events in their approach. Li et al.~\cite{liNeuromorphicDataAugmentation2022} applied image morphological transformations (rolling, cutout, shear, and rotation) to the event stream data, but they ignored the temporal properties of the event stream data.

In this study, we apply data augmentation based on sample mixing to event stream data for the first time and design an augmentation method based on the spatio-temporal dynamics and sparsity of event stream data: EventMix. Also, our experiments show that the proposed method achieves state-of-the-art on some event stream datasets.

\section{EventMix}

\subsection{Event Stream Representation}
\label{mix-intr}
The raw, direct event stream data from the DVS can be represented as:

\begin{equation}
      \epsilon = \{e_i\}^I_{i=i} = \{t_i, x_i, y_i, p_i\}_{i=1}^I
      \label{event_stream}
\end{equation}

Where $t$ denotes the timestamp of the event, $(x, y)$ denotes the position and $p$ denotes the polarity, with $+1$ and $-1$ indicating the increase and decrease of light, respectively.

In order to enable neural networks to process the event stream data quickly and efficiently, we convert the event data into frame-based, which is commonly used for the event data. Similar to the method of~\cite{gehrigEndtoEndLearningRepresentations2019}, we first split the event stream data with duration T on average. The bin size of each frame is $\Delta T$, and the events in $\Delta T$ are summed according to their polarities. Finally, we can get frame-based data with two channels and a length of $T/\Delta T$. The details are shown in Eq.~\ref{slice1}:

\begin{equation}
      E(c, x, y, p) = \sum_{e_i \in \epsilon} k(t_i - t_c, x_i - x, y_i - y, p_i - p)
      \label{slice1}
\end{equation}

where, $t_c$ indicates the start time of the $c$ th frame after splitting, $k(t, x, y, p)$ can be expressed as:

\begin{equation}
      k(t, x, y, p) = \delta(x, y, p) (t < \Delta T)
      \label{slice}
\end{equation}

Our EventMix shares a similar motivation with MixUp and CutMix, in that it mixes two samples and labels to generate the new samples and their corresponding labels.

\subsection{Improved 3D dynamic Masking}
\label{mix-mask}
Let $x $ denote the sample in the training set after the above transformation, then the mixed samples $\widetilde{x}$ can be obtained from the two training samples $(x_A, x_B)$   :
\begin{align}
      \widetilde{x} & = \textbf{M} \odot x_A + (\textbf{1} - \textbf{M}) \odot x_B
      \label{mix}
\end{align}

$\textbf{M} \in \{0, 1\}$ denotes a binary mask, and the original data in the mask is discarded and filled with the corresponding part of the other data. For the shape of the mask, the original CutMix only considers the $\textbf{M}$ in the spatial dimension and restricts the shape to be square. This unnecessary limitation on the data will make the models more biased toward the general features and is not conducive to the performance. Here, we remove the restriction that the mask must be square and fully consider the spatiotemporal characteristics of event data to extend the mask to 3 dimensions. We first generated a continuous 3D mask by random Gaussian Mixture Model and then generated a binary 3D mask by binarization:

\begin{equation}
      \textbf{M} = B(\sum_{k=1}^{K} \pi_k N(\textbf{X} \mid \mu_k, \Sigma_k), \lambda)
      \label{mask}
\end{equation}

$K$ denotes the number of components in the Gaussian Mixture Model, $\pi_k$, $\mu_k$, and $\Sigma_k$ denote the mixing coefficient, mean, and variance of the $k$ th component, respectively. The above variables are generated randomly. $\lambda$ denotes the proportion of the randomly generated fraction to be mixed with the original sample and is obtained by sampling from the $beta(1, 1)$ distribution. $B(\cdot, \lambda)$ represents the transformation of the real mask obtained by the random Gaussian Mixture Model into a binarized form:

\begin{equation}
      B(\textbf{X}, \lambda)_{t, i, j} = \\
      \begin{cases}
            1, & \textbf{X}_{t, i, j} < top(\lambda \ size(\textbf{X}), X) \\
            0, & otherwise
      \end{cases}
      \label{gate}
\end{equation}

\begin{figure}[htbp]
      \centering
      \includegraphics[width=12cm]{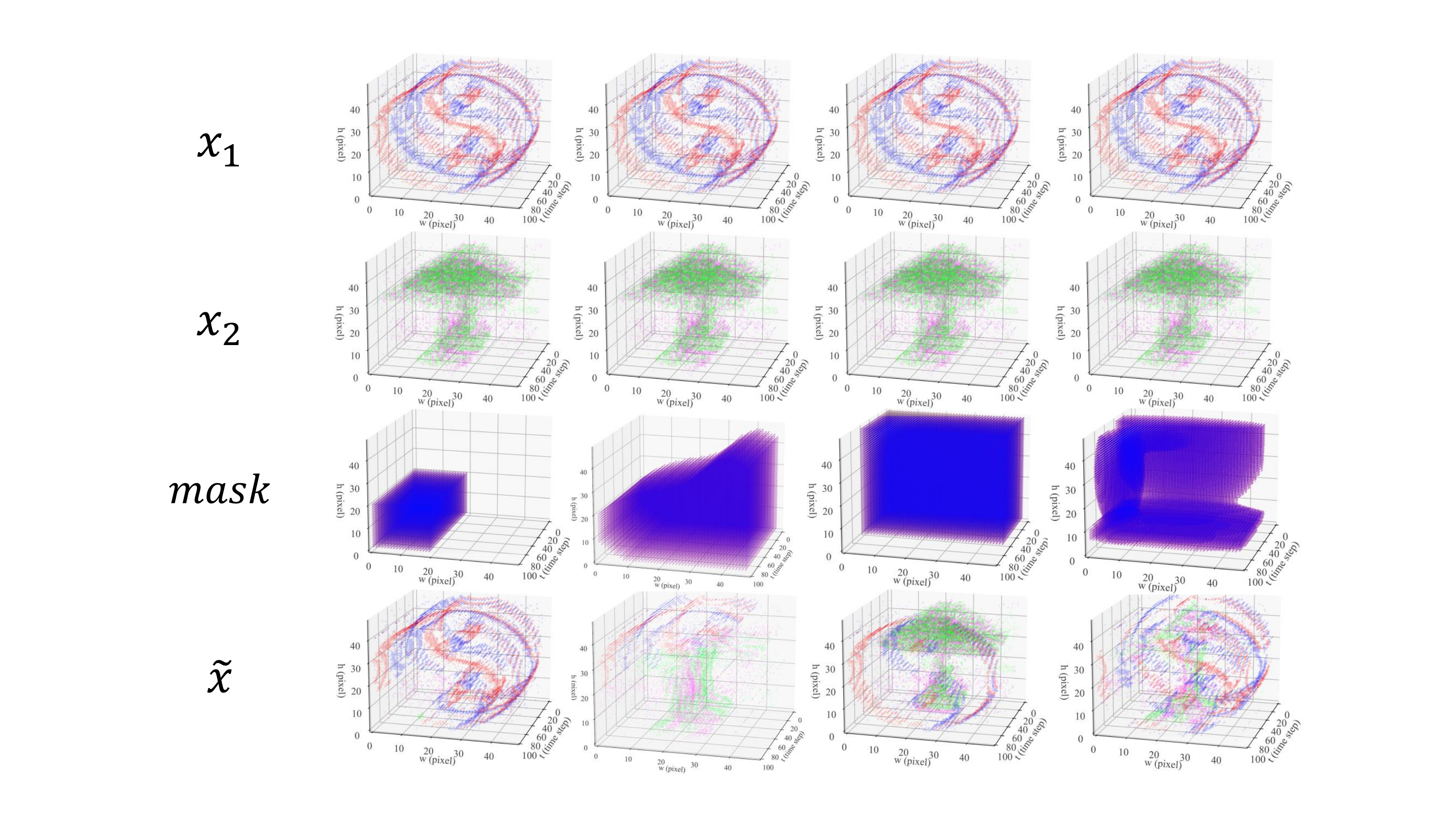}
      \caption{Some examples of data augmentation using different masks. From left to right, we show the use of square masks, spatial masks, temporal masks, and spatio-temporal masks.}
      \label{vis}
\end{figure}

$top(k, \textbf{X})$ denotes the value of the $k_{th}$largest number in $\textbf{X}$. As shown in Figure~\ref{vis}, By applying the above method, it is possible to generate diverse masks with temporal and spatial dimensions and achieve more mixing patterns of event stream data.

\subsection{Mixing of Event Labels}
\label{label-mix}
Let $y$ denote the labels, then the corresponding label $\widetilde{y}$ of the mixed sample in Eq.~\ref{mix} is shown as Eq.~\ref{mix2}:
\begin{align}
      \widetilde{y} & = \alpha y_A + (1 - \alpha) y_B
      \label{mix2}
\end{align}

\begin{figure}[htbp]
      \centering
      \includegraphics[width=11cm]{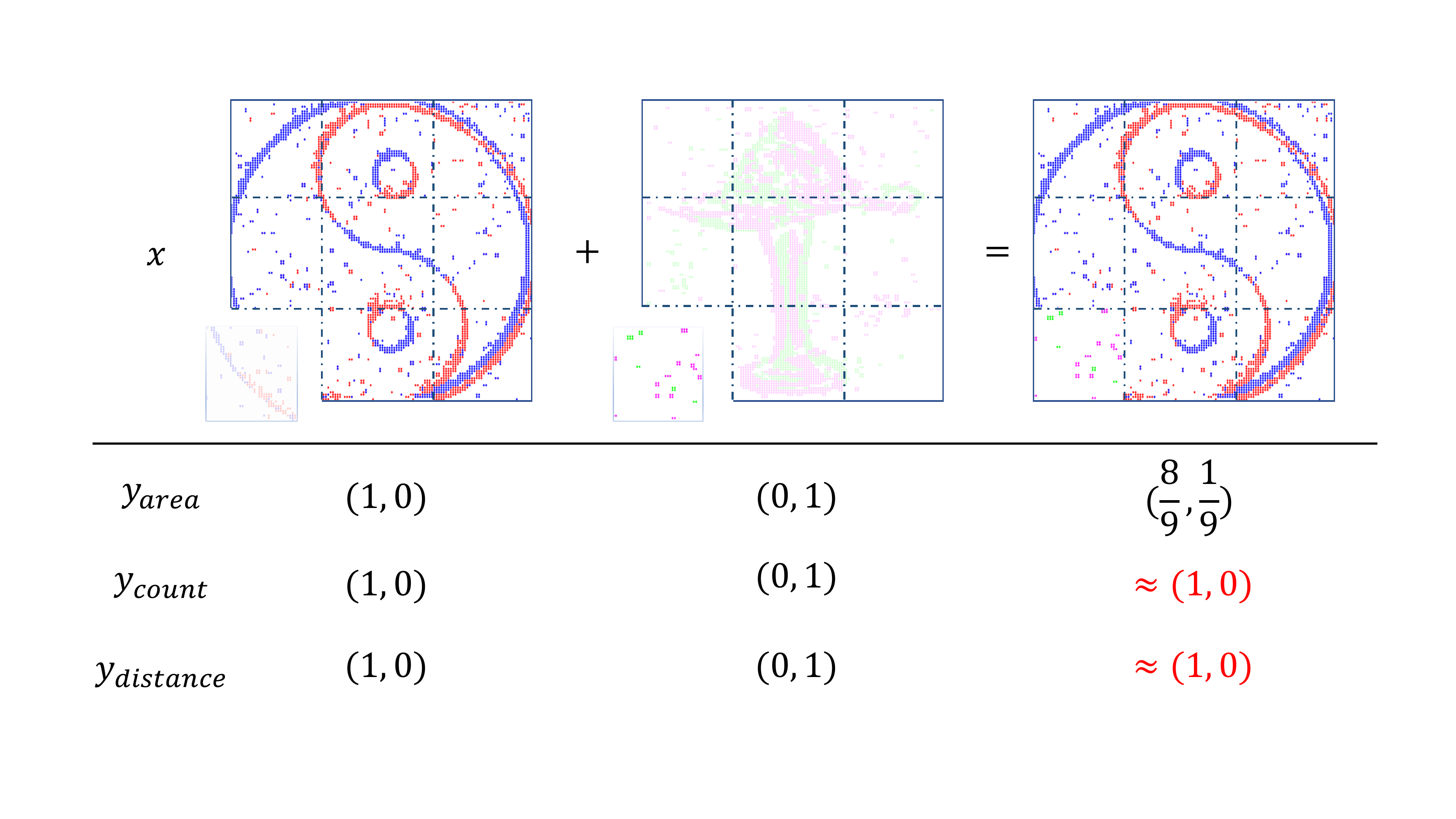}
      \caption{An example of different label weight assignment methods of event stream data. }
      \label{assignment}
\end{figure}

As shown in Figure~\ref{assignment}, unlike the image data, the event stream data consists of sparse events, and meaningful regions in event stream data can be easily distinguished. It is not suitable to use the area for label weight assignment. Therefore, we design two reasonable label mixing methods: based on the number of events and the relative distance of events.

$\textbf{Based on the number of events}$. Sample mixing means that the events in the subregion of the original data $x_A$ are deleted and filled with events from the same location of another event $x_B$. So a simple idea is to calculate the proportion of deleted events to $x_A$ and the proportion of added events to $x_B$, and calculate the mixed label weight $\alpha$:

\begin{equation}
      \alpha = \frac{\frac{\sum \textbf{M} \odot x_A}{\sum x_A}}{\frac{\sum \textbf{M} \odot x_A}{\sum x_A} + \frac{\sum (\textbf{1} - \textbf{M}) \odot x_B}{\sum x_B}}
      \label{count}
\end{equation}

Where $\sum{\cdot}$ means summing over all elements of the tensor. Based on Eq.~\ref{count}, the weights of the mixed labels can be calculated and combined with Eq.~\ref{mix} to obtain a representation of the mixed labels based on the proportion of the number of events in the mixed region.

$\textbf{Based on the relative distance}$. Directly calculating the number of events in the mixed region solves the problem of biased label weight assignment when the mixed region is background. However, this approach only considers the number of events in the mixing region but ignores the distribution of events. Considering the sparse data of event streams and using only binary representation, we design a way to calculate the distance of event streams and use it as the basis for label weight assignment:

\begin{equation}
      E(x_A, x_B) = e(f(x_A), f(x_B))
      \label{distance}
\end{equation}

Where $e(\cdot, \cdot)$ indicates how to calculate the distance between two event streams. In this study, we use the mean square error (MSE) to represent the distance between event streams. The $f(\cdot)$ denotes the preprocessing of the event streams. Since calculating the difference of corresponding pixels directly is a strict metric, considering that the event streams have translation invariance, we first process the event stream using the average pooling operation and then calculate the distance.

\begin{equation}
      \alpha = \frac{E(x_B, \widetilde{x})^2}{E(x_A, \widetilde{x})^2 + E(x_B, \widetilde{x})^2}
\end{equation}

According to the characteristics of event stream data, we design the data augmentation strategy based on sample mixing. More diverse event stream mixing is achieved, as well as more reasonable label weight assignment rules.

\section{Experiments}
In this chapter, we will conduct experiments on several datasets to evaluate the effectiveness of our EventMix. The experiments are based on the PyTorch framework~\cite{paszkeAutomaticDifferentiationPyTorch} on NVIDIA A100 GPU with several datasets to verify our method. We use the AdamW~\citep{loshchilovDecoupledWeightDecay2019a} as the optimizer, the learning rate $lr$ is set with $1\times 10^{-3}$. The training epochs are set at 500.

Random cropping, random flipping, and random rotation are used on the training samples before other mixing-based augmentation operations. If not mentioned, Eq.~\ref{distance} is used as the default method. We conduct experiments on ANNs and SNNs with PLIF \cite{fangIncorporatingLearnableMembrane2021} neurons on several network structures. Since in spiking neural networks, information is passed between layers in the form of spikes, which leads to low information density and tends to degrade performance. Therefore, we use the same method as~\cite{heIdentityMappingsDeep2016} to preactivate each block of ResNet and pass the membrane potential between modules with identity connectivity, which ensures more information transfer without increasing the computation of the network. We compare the performance of our proposed EventMix with CutMix and MixUp for event stream data augmentation with the same neural network model and hyperparameter settings.

\subsection{Implementation Details}
\label{details}
We validated EventMix on the following four DVS datasets:

$\textbf{DVS-CIFAR10}$~\cite{liCIFAR10DVSEventStreamDataset2017}. DVS-CIFAR10 is a neuromorphic version converted from the CIFAR10 dataset~\cite{krizhevskyLearningMultipleLayers}. 10,000 frame-based images were transformed with DVS into 10,000 event streams. All event streams are sliced by Eq.~\ref{slice} before training or validation, and in the same way as Li et al.~\cite{liNeuromorphicDataAugmentation2022} we slice the event stream data into 10 time steps. Before data augmentation, we resize all event streams to $48 \times 48$ in the spatial dimension. Since this dataset does not divide the training and valid sets, we divided the training and valid sets by $9:1$.

$\textbf{N-Caltech101}$~\cite{orchardConvertingStaticImage2015}. N-Caltech101 is the event stream version of Caltech101~\cite{fei-feiLearningGenerativeVisual2004}. The images in Caltech101 are displayed on the LCD monitor and captured by an automatically moving event camera. After this transformation, the N-Caltech101 dataset is obtained. Since there is a difference in the number of samples of different categories in N-Caltech101, we resample the training set according to the ratio of the number of different categories. As in DVS-CIFAR10, we resize the event stream to $48 \times 48$, divide it into $10$ time steps, and divide the training and validation sets by $9:1$.

$\textbf{N-CARS}$~\cite{sironiHATSHistogramsAveraged2018} is a real-world event stream dataset acquired by an event camera placed behind the front windshield of a car in a real-world road environment. The samples in this data are divided into two categories: background and vehicle, with $15422$ training samples and $8607$ test samples.

$\textbf{DVS-Gesture}$~\cite{amirLowPowerFully2017} is a real-world gesture recognition dataset collected by the DVS camera. It contains $1342$ different packs of $11$ categories collected on $29$ individuals. We used the same preprocessing approach as Gregor et al.~\cite{lenzTonicEventbasedDatasets2021} and the training and validation sets were divided by $8:2$.

\subsection{Comparison with Existing Literature}
\label{existing_literature}
To illustrate the superiority of our EventMix, we show experimental results against other state-of-the-art algorithms. As shown in Table~\ref{comparison}, for those four datasets, both ANN and SNN, our EventMix brings a significant improvement. Especially for the DVS-Gesture dataset, EventMix improves by nearly 12\% on SNNs and 5\% on ANNs. Meanwhile, on the N-Caltech 101 dataset, our model outperforms Neuromorphic data augmentation by nearly 11\%, and even compared to EventDrop, our model outperforms by nearly 4\%. EventDrop just randomly drops the event, which destroys the distribution of the original event data but does not consider the change of the label. In addition to mixing samples, our EventMix fully considers the spatiotemporal characteristics of event data and aligns more accurate and reasonable labels to the mixed data, which greatly improves the performance of the algorithm.

\begin{table}[h]
      \resizebox{\textwidth}{!}{
            \centering
            \begin{threeparttable}[b]
                  \caption{Comparison of classification results with other methods.}
                  \begin{tabular}{cccccc}
                        \toprule[1pt]
                        Method                                                     & Model        & DVS-CIFAR10      & N-Caltech101   & N-CARS         & DVS-Gesture    \\
                        \hline
                        BNTT~\cite{kimRevisitingBatchNormalization2021}            & 6-layer CNN  & 63.2             & -              & -              & -              \\
                        Rollout~\cite{kugeleEfficientProcessingSpatioTemporal2020} & VGG16        & 66.5             & -              & 94.07          & 95.68          \\
                        SALT~\cite{kimOptimizingDeeperSpiking2021}                 & VGG11        & 67.1             & 55.0           & -              & -              \\
                        LIAF-Net~\cite{wuLIAFNetLeakyIntegrate2021}                & VGGlike      & 70.4             & -              & -              & -              \\
                        tdBN~\cite{zhengGoingDeeperDirectlyTrained2020}            & ResNet19     & 67.8             & -              & -              & -              \\
                        PLIF~\cite{fangIncorporatingLearnableMembrane2021}         & VGG11        & 74.8             & -              & -              & -              \\
                        EST~\cite{gehrigEndtoEndLearningRepresentations2019}       & ResNet34-ANN & -                & 81.7           & 92.5           & -              \\
                        \midrule[.5pt]
                        w/o NDA~\cite{liNeuromorphicDataAugmentation2022}          & ResNet19     & 67.9             & 62.8           & 82.4           & -              \\
                        w/. NDA~\cite{liNeuromorphicDataAugmentation2022}          & ResNet19     & 78.0             & 78.6           & 87.2           & -              \\
                        \midrule[.5pt]
                        w/o EventDrop~\cite{guEventDropDataAugmentation}           & ResNet34-ANN & -                & 83.91          & 91.03          & -              \\
                        w/. EventDrop~\cite{guEventDropDataAugmentation}           & ResNet34-ANN & -                & 85.15          & 95.50          & -              \\
                        \midrule[.5pt]
                        Baseline                                                   & Resnet18     & 79.23            & 75.25          & 95.94          & 84.76          \\
                        EventMix                                                   & Resnet18     & $\textbf{81.45}$ & \textbf{79.47} & \textbf{96.29} & \textbf{96.75} \\
                        Baseline                                                   & Resnet34-ANN & 81.13            & 84.51          & 95.36          & 86.33          \\
                        EventMix                                                   & Resnet34-ANN & \textbf{85.60}   & \textbf{89.20} & \textbf{96.54} & \textbf{91.80} \\
                        \bottomrule[1pt]
                  \end{tabular}
                  \label{comparison}
            \end{threeparttable}
      }
\end{table}

\subsection{Comparison with Existing Mixing-based Methods}
In \ref{existing_literature}, we show a comparison of our proposed EventMix strategy with some extant literature and confirm that our approach is able to achieve advanced performance on event stream data classification tasks. To further illustrate the efficiency and effectiveness of the EventMix strategy, we compare it with some advanced sample mixing-based data augmentation strategies applied directly to event stream data. We use MixUp and CutMix to compare with our proposed method. In order to apply the above two methods to 3D event data, we make a simple extension of them. For MixUp, we directly mix events from different event data with different weights, as for CutMix we generate a two-dimensional mask and then repeat it in the temporal dimension to generate a 3D mask. The results are shown in Table~\ref{mixing-based}.

\begin{table}[h]
      \centering
      \caption{Comparison of mixing-based data augmentation with ANN.}
      \scalebox{0.9}{
            \begin{tabular}{cccccc}
                  \toprule[1pt]
                  \multirow{2}{*}{Model}    & \multirow{2}{*}{Method} & \multicolumn{3}{c}{Accuracy (Improvement)}                                                           \\
                  \cline{3-5}
                                            &                         & DVS-CIFAR10                                & N-Caltech101               & DVS-Gesture                \\
                  \hline
                  \multirow{4}{*}{ResNet34} & Baseline                & 81.13$_{+0.00}$                            & 84.51$_{+0.00}$            & 86.33$_{+0.00}$            \\
                                            & MixUp                   & 82.93$_{+1.80}$                            & 83.85$_{+2.72}$            & 89.06$_{+2.73}$            \\
                                            & CutMix                  & 83.15$_{+2.02}$                            & 86.84$_{+2.33}$            & 87.11$_{+0.78}$            \\
                                            & EventMix                & $\textbf{85.60}$$_{+4.47}$                 & $\textbf{89.20}$$_{+4.69}$ & $\textbf{91.80}$$_{+5.47}$ \\
                              \hline
                  \multirow{4}{*}{ResNet18} & Baseline                & 80.24$_{+1.79}$                            & 82.03$_{+1.79}$            & 85.55$_{+0.00}$            \\
                                            & MixUp                   & 83.15$_{+2.91}$                            & 84.38$_{+4.14}$            & 86.72$_{+1.17}$            \\
                                            & CutMix                  & 82.16$_{+1.92}$                            & 81.58$_{+1.34}$            & 86.33$_{+0.78}$            \\
                                            & EventMix                & $\textbf{84.38}$$_{+4.14}$                 & $\textbf{84.71}$$_{+4.47}$ & $\textbf{89.45}$$_{+3.90}$ \\
                        \hline
                  \multirow{4}{*}{MobileV2} & Baseline                & 79.46$_{+0.00}$                            & 76.30$_{+0.00}$            & 78.91$_{+0.00}$            \\
                                            & MixUp                   & 82.37$_{+2.91}$                            & 83.29$_{+6.99}$            & 81.25$_{+2.34}$            \\
                                            & CutMix                  & 83.37$_{+3.91}$                            & 81.77$_{+5.47}$            & 82.03$_{+3.12}$            \\
                                            & EventMix                & $\textbf{83.70}$$_{+4.24}$                 & $\textbf{83.85}$$_{+7.55}$ & $\textbf{82.93}$$_{+4.02}$ \\
                  \bottomrule[1pt]
            \end{tabular}
      }
      \label{mixing-based}
\end{table}

In Table~\ref{mixing-based}, we compare our proposed EventMix strategy with some mixing-based data augmentation strategies designed for image data using the same ANN structure under the same hyperparameter settings. MixUp mixes all events in two event streams according to their weights, and since event data is very sparse and has simple semantics, mixing samples in this way often causes ambiguity. CutMix can avoid the above problem, but this approach restricts the data mask to rectangles and ignores the temporal dimensionality of the event data. In addition, the label weight assignment of these two methods does not consider the characteristics of the event data. Therefore, the data augmentation strategies MixUp and CutMix designed for image data can only provide limited performance improvement for event stream data. By carefully considering the characteristics of event stream data, our proposed EventMix strategy can improve the model performance by $2.45\%$ on the DVS-CIFAR10 dataset, $2.36\%$ on the N-Caltech101 dataset, and $4.69\%$ on DVS-Gesture dataset when using the ResNet34-ANN model compared to the CutMix strategy.

\subsection{Component Analysis}

Our proposed EventMix strategy redesigns the sample mixing-based data augmentation strategy from two aspects: mask generation and label weight assignment. On the one hand, we design a 3D mask generation method that can mix different samples in time and space dimensions according to the characteristics of event stream data and reduce the impact of monotonous mask shapes on the model's generalization ability. On the other hand, based on the sparse data characteristics of event streams, we first assume that each event in the event stream has the same contribution to the corresponding label and design two label mixing methods based on the number of events and the relative distance of the event streams. Therefore, we then compare different masking approaches and different label weight assignment approaches and further illustrate the effectiveness and efficiency of our proposed EventMix strategy.

\begin{table}[h]
      \centering
      \caption{Component Analysis with ResNet34-ANN on DVS-CIFAR10.}
      \begin{tabular}{ccccc}
            \toprule[1pt]
            Method   & Area            & Count           & Distance                   \\
            \hline
            Square   & 82.59$_{+0.00}$ & 82.14$_{-0.45}$ & 82.37$_{-0.22}$            \\
            Spatial  & 82.70$_{+0.11}$ & 82.59$_{0.00}$  & 83.48$_{+0.89}$            \\
            Temporal & 84.15$_{+1.56}$ & 84.38$_{+1.79}$ & 83.37$_{+0.78}$            \\
            S \& T   & 84.04$_{+1.45}$ & 85.16$_{+2.57}$ & $\textbf{85.60}$$_{+3.01}$ \\
            \bottomrule[1pt]
      \end{tabular}
      \label{component}
\end{table}

Table~\ref{component} lists the performance contributions of the different components of our proposed EventMix strategy. The use of square masks and sample mixing in the spatial dimension is the result of directly applying CutMix to the event stream data. From Table~\ref{component}, it can be seen that the mixing of event streams in the temporal dimension is essential and can improve the accuracy of the model by almost $1\%$. Using the random Gaussian Mixture Model to generate 3D masks can enrich the pattern of sample mixing and further improve the generalization ability of the model. Our proposed event stream data label mixing methods can more accurately assign labels to the mixed samples and improve the model performance than the original area-based mixing method. Combining these two strategies can generate more diverse event stream data mixing patterns, assign mixed sample labels more rationally, and achieve a $3.01\%$ performance improvement over the original data augmentation strategy.

\subsection{Effect of the Amount of Training Sample}

\begin{figure}[htbp]
      \centering
      \includegraphics[width=8.5cm]{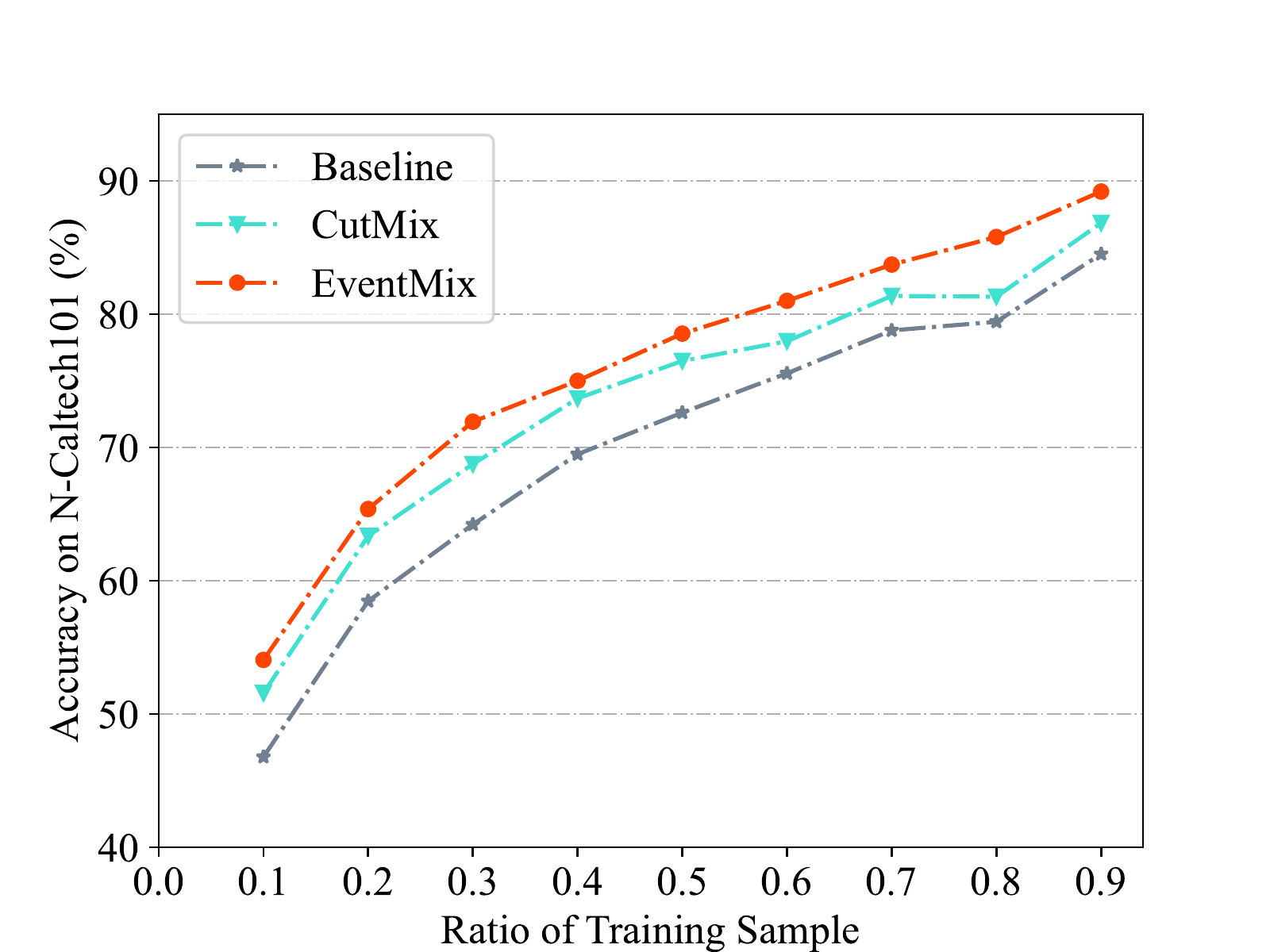}
      \caption{Comparison of EventMix with CutMix and baseline when using different number of samples as training set.}
      \label{amount}
\end{figure}

To further demonstrate the effectiveness of the EventMix data augmentation strategy in extreme cases, we compared the performance of the EventMix strategy under different training sample sizes with some other data augmentation strategies.

Figure~\ref{amount} shows the performance of different data augmentation algorithms when using $10\%$ - $90\%$ of the total N-Caltech101 dataset as the training data. Our proposed data augmentation strategy guarantees $54.06\%$ validation accuracy with only $10\%$ of the data as the training set, a $7.28\%$ improvement over the conventional morphological transformation strategy. It maintains a significant advantage over the baseline for different training set sizes.

\section{Conclusion}

Since the event data is small and difficult to obtain, this paper designs an efficient data augmentation strategy EventMix for the event data. EventMix has removed the limitation of the square mask in the CutMix and designs a three-dimensional mask according to the spatiotemporal characteristics of event data. Also, a more reasonable label assignment is designed for the mixed sample. We have tested on multiple event datasets, and the experimental results show that our EventMix can significantly improve the performance of ANNs and SNNs on the event-based dataset. For SNNs, our EventMix has reached state-of-the-art performance on DVS-CIFAR10, N-Caltech101, N-CARS, and DVS-Gesture datasets.

\bibliography{refs}
\bibliographystyle{icml2021}

\end{document}